\documentclass[preprint]{elsarticle}
\usepackage{amsmath,amssymb,amsfonts,amsthm,calc}
\usepackage{eucal} 

\usepackage{caption}
\usepackage{subcaption}
\usepackage[all]{xy}
\usepackage{graphicx}
\usepackage{enumerate}

\newcounter{ecount}
\newcommand{\R}{\mathbb{R}}

\begin{document}

\begin{frontmatter}

\title{Classification of Hepatic Lesions using the Matching Metric}
\author[EE]{Aaron ~Adcock\corref{cor2}}
\ead{aadcock@stanford.edu}

\author[rad]{Daniel ~Rubin}  
\ead{dlrubin@stanford.edu}

\author[math]{Gunnar ~Carlsson}
\ead{gunnar@math.stanford.edu}

\cortext[cor2]{Principal corresponding author, Tel: 1-740-405-5052 Fax: 1-650-725-4066 }

\address[EE]{Dept. of Electrical Engineering, 350 Serra Mall Stanford, CA 94305}

\address[rad]{Dept. of Radiology, Richard M. Lucas Center, P285, Stanford, CA 94305}

\address[math]{Dept. of Mathematics, Bldg 380, Stanford, CA 94305}

\begin{abstract}
In this paper we present a methodology of classifying hepatic (liver) lesions using multidimensional persistent homology, the matching metric (also called the bottleneck distance), and a support vector machine.  We present our classification results on a dataset of 132 lesions that have been outlined and annotated by radiologists. We find that topological features are useful in the classification of hepatic lesions.  We also find that two-dimensional persistent homology outperforms one-dimensional persistent homology in this application.
\end{abstract}


\end{frontmatter}

\section{Introduction}
\label{sec:data}

Medical imaging technology allows doctors access to portions of the human body which are visually inaccessible to the human eye.  Often inspecting these medical images is a labor intensive process performed by diagnostic radiologists.  The accuracy of the radiologist is obtained through training and experience \cite{autoretrieve} but even with extensive training and experience there are variations in interpretations and accuracy among radiologists \cite{lung,radvar}.  Despite an increasing emphasis on evidence-based medicine and improved imaging techniques, quantitative `gold-standards' and clear guidelines for a radiologist's role in quantitative measurements remain elusive \cite{quant}.  Image processing provides a way of both automating portions of the examination as well as providing standard tools for radiologists to use when reading an image.  The qualitative nature of many radiological observations suggests that topological features may be useful in the classification and interpretation of medical images.

\begin{figure}[!h]
\begin{center}
\includegraphics[width=2.5in]{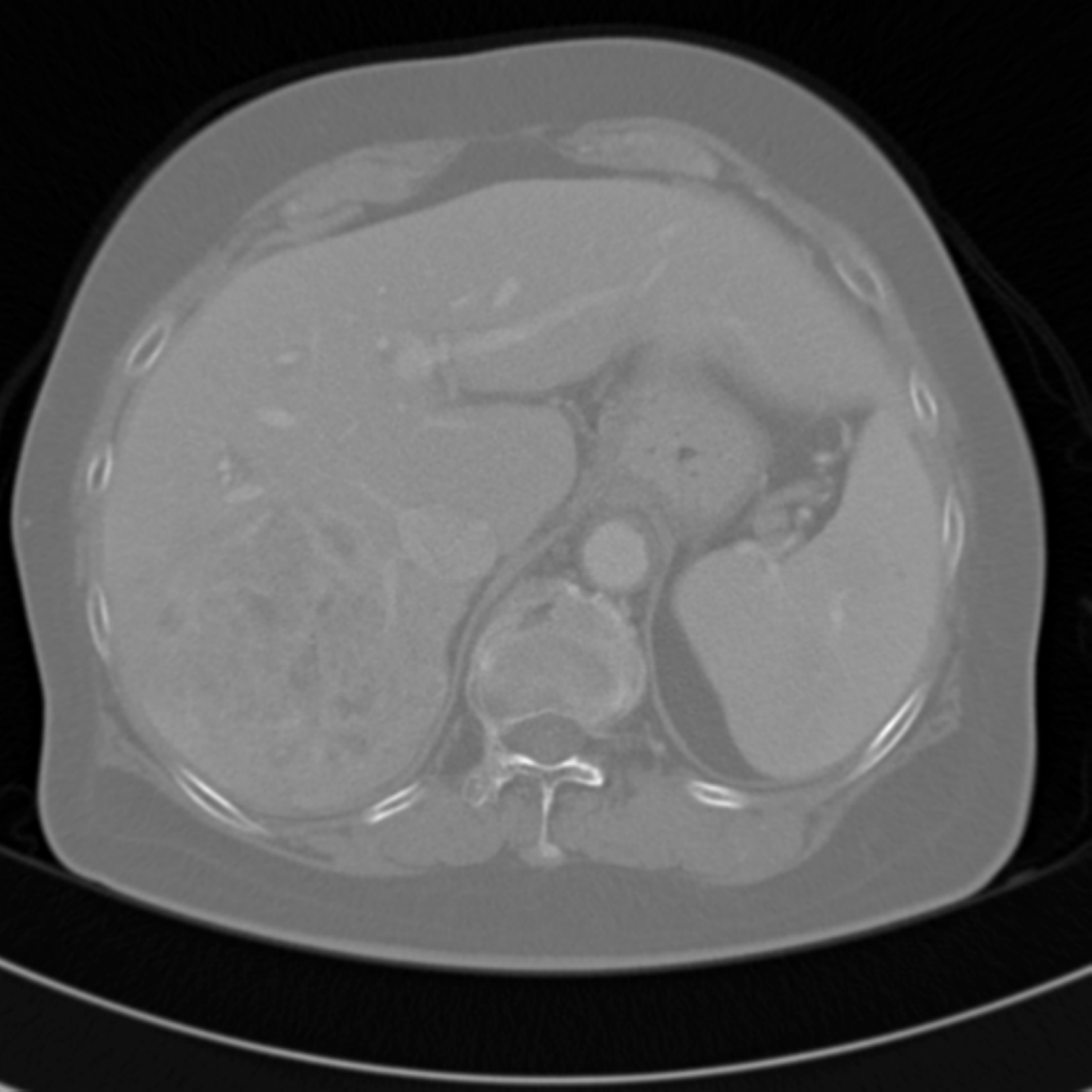}
\caption{Abdominal CT Scan}
\label{fig:ct}
\end{center}
\end{figure}

In this paper, we explore automatic classification methods of computed tomography (CT) scans of hepatic (liver) lesions.  We have a dataset of CT scans of 132 hepatic lesions along with an outline, diagnosis, and semantic descriptors of the lesion provided by a radiologist.  There are nine lesion types represented in the data, with the vast majority of the lesions (90 lesions) evenly split between cysts and metastases, followed by hemangiomas (18 lesions), hepatocellular carcinomas (HCC, 11 lesions), focal nodules (5 lesions), abscesses (3 lesions), neuroendocrine neoplasms (NeN, 3 lesions), a single laceration and a single fat deposit.

\begin{figure}[!h]
\begin{center}
\begin{subfigure}{1.5in}
\begin{center}
\includegraphics[width=.85in]{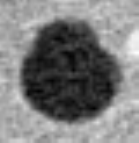}
\caption{Cyst}
\end{center}
\end{subfigure}
\begin{subfigure}{1.5in}
\begin{center}
\includegraphics[width=.85in]{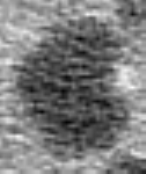}
\caption{Metastasis}
\end{center}
\end{subfigure}
\begin{subfigure}{1.5in}
\begin{center}
\includegraphics[width=.85in]{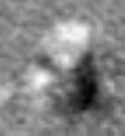}
\caption{Hemangioma}
\end{center}
\end{subfigure}
\begin{subfigure}{1.5in}
\begin{center}
\includegraphics[width=.85in]{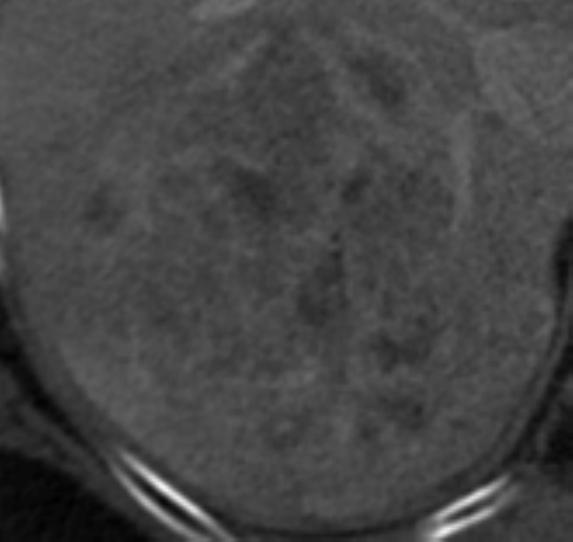}
\caption{HCC}
\end{center}
\end{subfigure}
\begin{subfigure}{1.5in}
\begin{center}
\includegraphics[width=.85in]{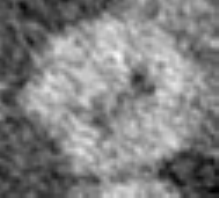}
\caption{Focal Nodule}
\end{center}
\end{subfigure}
\begin{subfigure}{1.5in}
\begin{center}
\includegraphics[width=.85in]{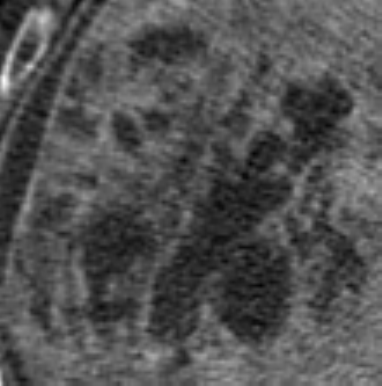}
\caption{Abscess}
\end{center}
\end{subfigure}
\begin{subfigure}{1.5in}
\begin{center}
\includegraphics[width=.85in]{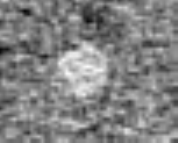}
\caption{NeN}
\end{center}
\end{subfigure}
\begin{subfigure}{1.5in}
\begin{center}
\includegraphics[width=.85in]{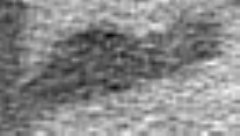}
\caption{Laceration}
\end{center}
\end{subfigure}
\begin{subfigure}{1.5in}
\begin{center}
\includegraphics[width=.85in]{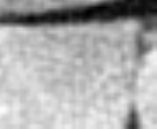}
\caption{Fat Deposit}
\end{center}
\end{subfigure}

\caption{Lesion Diagnoses}
\label{fig:diagnose}
\end{center}
\end{figure}

\begin{figure}[!h]
\begin{center}
\begin{subfigure}{1.5in}
\begin{center}
\includegraphics[width=1in]{40_abscess_lesion.pdf}
\end{center}
\end{subfigure}
\begin{subfigure}{1.5in}
\begin{center}
\includegraphics[width=1in]{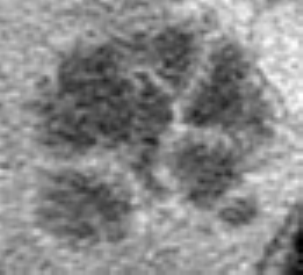}
\end{center}
\end{subfigure}
\begin{subfigure}{1.5in}
\begin{center}
\includegraphics[width=1in]{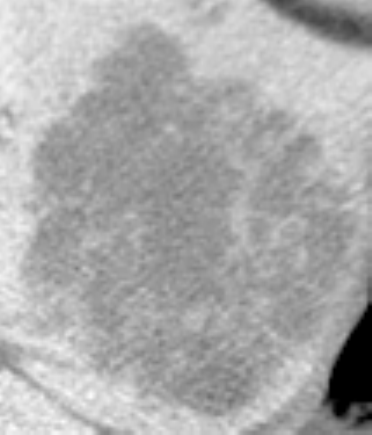}
\end{center}
\end{subfigure}
\end{center}
\caption{Abscesses}
\label{fig:absc}
\end{figure}

\begin{figure}[!hbt]
\begin{center}
\begin{subfigure}{1.5in}
\begin{center}
\includegraphics[width=1in]{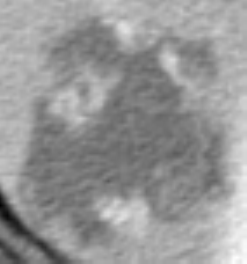}
\end{center}
\end{subfigure}
\begin{subfigure}{1.5in}
\begin{center}
\includegraphics[width=1in]{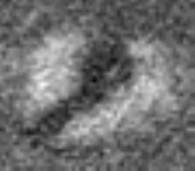}
\end{center}
\end{subfigure}
\begin{subfigure}{1.5in}
\begin{center}
\includegraphics[width=1in]{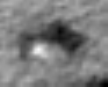}
\end{center}
\end{subfigure}

\caption{Hemangiomas}
\label{fig:hem}
\end{center}
\end{figure}

It has been demonstrated that semantic features are useful for classification in hepatic lesions \cite{autoretrieve}.  This indicates that visually identifiable structures exist within the lesions, but it has been difficult finding quantitative methods of defining these structures.  For example, consider the six images in Figures \ref{fig:absc} and \ref{fig:hem}.  The first three show the abscesses contained in our dataset.  The second three show hemangiomas (deformations of blood vessels).  The abscesses present what is called `cluster of grapes' morphology.  But the arrangements of this structure (the clusters of grapes) are very different in each lesion.  Similarly, the hemangiomas show the characteristic large dark central region with dense white regions on the outer edge of the lesion.  Yet, the hemangiomas lack a rotational orientation, different numbers of the two region types exist and the formations vary in size and shape.  The qualitative nature of these observations has made it difficult to find quantitative measures of the structures.


\subsection{Prior Work}

As mentioned above, semantic features have been one successful method for classifying the liver lesions \cite{autoretrieve}.  This has led to preliminary investigations into using computational features to predict semantic features, which can be used to classify the lesions \cite{semantic}.  Additionally, computational features have shown some success in directly classifying liver lesions \cite{semantic, autoretrieve,  gabor}. Most of these studies use a large number of various types of features (intensity histograms, wavelets, boundary features, etc) to classify the lesions.  Shape descriptors for liver lesions have also been investigated and found to work well in retrieving similar lesions \cite{livshape}.

Persistent homology \cite{topdata, persistence} is an approach to extending the notion of shape to point clouds  or finite metric spaces, which has been developed over the last 10-15 years.  It has two directions of application, one which gives understanding of the overall organization of data sets (see e.g. \cite{range, patches, parapatch}),  and a second which applies to data sets consisting of data points which themselves have complex structure, such as databases of images or of chemical structures.  This paper approaches radiological images from the second point of view.  Persistent homology depends on a family of simplicial complexes parametrized by a real variable.  In many applications, including the ones in the first direction, this parameter is simply a scale variable, which measures distances between points.  In applications in the second direction, however, one uses families depending on other parameters, and often uses multiple persistence invariants in conjunction with each other to obtain useful information (see \cite{shapes, curves, melanomas}).  In this paper we will find an appropriate set of parameters and functions of those parameters for classification problems arising in the radiology of liver lesions.

\subsection{Our Work}

In this paper we present a framework for computing persistent homology on images.  This framework is flexible and allows the user to tailor the filtrations to the application.  We demonstrate this by applying this framework to classifying our set of hepatic lesions and using the bottleneck distance to compare lesions.  We find that our method gets comparable results to existing methods and our results demonstrate the possibilities of tailoring multidimensional persistence to a specific application.  Additionally, since we use an `off-the-shelf' implementation of the support vector machine to perform the final classification, our results demonstrate that this framework can be easily integrated with existing classification techniques.

\section{Theory}
\label{sec:back}
The work presented in this paper is an application of computational topology to medical image processing.  As such, the theoretical material will be presented in a compact, informal manner following \cite{shapes, curves}.  Those interested in the details behind the theory are encouraged to read the associated references.

\subsection{Filtrations \& Simplicial Complexes}
Let $V$ be a set of points.  A \emph{k-simplex} $S \subseteq V$ of size $k+1$.  A geometric realization of a simplex consists of the convex hull of $k+1$ affinely independent points in $\R^d, d \ge k$.  A 0-simplex is a point or \emph{vertex}, a 1-simplex is an \emph{edge}, a 2-simplex is a \emph{triangle}, and a 3-simplex is a \emph{tetrahedron}.  Higher dimensional simplices are difficult to visualize and thus less familiar to our experience.  A \emph{simplicial complex} on $V$ is a set $K$ of simplices on $V$ such that the following holds: If the simplex $\sigma \in K$, and the simplex $\tau \subset \sigma$ then $\tau \in K$.  A \emph{subcomplex} of $K$ is a simplicial complex $L \subseteq K$.  A \emph{filtration} of a complex $K$ is a nested sequence of complexes $\emptyset = K_{min} \subseteq K_{min+1} \subseteq \dots \subseteq K_{max} = K$.  We call $K$ a \emph{filtered complex}.  

\subsection{Persistent Homology}
\emph{Homology} is an algebraic invariant that counts some of the topological invariants of a complex in terms of its \emph{Betti numbers}, $\beta_i$.  Specifically, $\beta_0$ counts the number of connected components of $X$, $\beta_1$ counts the number of tunnels through $X$ (two-dimensional empty space enclosed by a one-dimensional curve), and $\beta_2$ counts the number of voids in $X$ (three-dimensional empty space enclosed by a two-dimensional surface).

Consider the filtered complex, a nested sequence of complexes, mentioned above.  We can track the topological changes that occur in this sequence via \emph{persistent homology}, an algebraic invariant which tracks the birth (appearance) and death (disappearance) of topological attributes (Betti numbers) during the evolution of this sequence.  The birth ($a$) and death ($b$) of a feature can be represented using the interval $[a,b] \subset \R$.  Counting the number of features in existence at a point $i$ (intervals which contain $i$) in the persistent homology of the filtered complex gives the homology of $K_i$.  Generally, the longer a feature's lifetime, the more important the feature is considered to be.  Features can have infinite lifetimes if they appear in all complexes after a given point in the sequence.

These concepts can be extended to multidimensional filtrations \cite{multi}. For example, in a two-dimensional filtration we have a sequence of simplicial complexes such that $K_{i_1,i_2} \subseteq K_{j_1,j_2}$ if $i_1 \le j_1$ or $i_2 \le j_2$.  Now each feature is represented as a two-dimensional `sheet' belonging to $\R^2$.  Except where noted, it can be assumed we are referring to one-dimensional filtrations.

\subsection{A metric on barcodes}
\label{sec:barcodes}
By computing the persistent homology of a filtered complex, we obtain a descriptor of the complex in the form of a finite multi-set of intervals, called a \emph{barcode}. Thus, the barcode is useful both as a data structure for storing the results of the computation of the persistent homology of a filtered complex and as a visual representation of the persistent homology.  A quasi-metric $D$, which we define as a metric which can take infinite values,  can be defined over the collection of all barcodes allowing us to compare complexes using this quasi-metric on the space of barcodes. We follow \cite{curves} in defining $D$.

Let $B_1$ and $B_2$ be two barcodes or finite multi-sets of intervals.  For two intervals, $I$ and $J$, we define their dissimilarity $\delta (I,J)$ to be their symmetric difference: $\delta(I,J) = \mu(I \cup J - I \cap J)$, where $\mu$ denotes the one-dimensional measure.  Note that $I,J$ can be infinite intervals and consequently $\delta (I,J)$ can be infinite.  A \emph{matching} $M$ on $B_1$ and $B_2$ is a set $M(B_1, B_2) \subseteq B_1 \times B_2 = \{ (I,J) | I \in B_1, J \in B_2\} $, where each interval occurs in at most one pair $(I,J)$.  Note that in general, $M$ will not have matched all intervals from $B_1$ and $B_2$.  Let $N$ be the set of unmatched intervals.  We can now defined $D_M(B_1, B_2)$, or the distance relative to $M$, as 

\begin{equation*}
D_M(B_1,B_2) = \sum_{(I,J) \in M)} \delta(I,J) + \sum_{L \in N} \mu (L)
\end{equation*}
We define the quasi-metric $D(B_1,B_2)$ as the best possible matching between $B_1,B_2$:

\begin{equation*}
D(B_1,B_2) = \min_{M} D_M(B_1,B_2)
\end{equation*}
This problem can be recast \emph{maximum weight bipartite matching problem} and solved using the \emph{Hungarian} algorithm.  See \cite{curves} for details.  We will abuse nomenclature slightly and refer to $D$ as the \emph{matching metric}.

\section{Calculations}
Each point in our dataset is an image, a two-dimensional collection of pixels (grayscale or intensity values) laid out on a grid with a set of contiguous pixels marked as lesion tissue.  To use computational topology to analyze each image, we need a method of forming a filtered complex from an image.  We can then use the theory outlined in Section \ref{sec:back} to create a barcode for each image and then use the matching metric to compare various images.

\subsection{Forming a Simplicial Complex from an Image}
\label{sec:simp}

Given a two-dimensional image $I$, we begin with the empty complex $K$.  We then assign a vertex to each pixel in $I$ and add each of these vertices (0-simplices) to $K$.  We then form 1-simplices from these vertices if the associated pixels are adjacent in $I$ (we treat diagonal pixels as adjacent).  We then add 2-simplices on the vertices where 3 pixels are mutually adjacent.  This forms a very regular simplicial complex, a mesh, with the only variations between images being the boundary shape of $I$.  As we are interested only in the hepatic lesions identified in the image by the radiologist, we define $I$ to be the region of pixels contained within the lesion outline, plus a border of healthy tissue around the edge of the lesion.  We set the width of this border to 5 pixels.  We keep this border because it is useful to have some healthy tissue included in the filtered complex for comparison with the lesion tissue.

\subsection{Image Filtrations}

After constructing the simplicial complex $K$ on the image $I$, we now want to define a filtration on $K$.  A natural approach is to assign a value to each vertex.  We can represent this as a function $f: V \rightarrow \R$, where $V$ is the vertex set of $K$.  Let $f_{min}$ and $f_{max}$ denote the minimum and maximum values obtained by $f$ on $V$.  We construct $K_i$ by including any simplex $S \in K$ with the property $\forall v \in S, f(v) \le i$.    
\begin{equation}
\label{eq:imfilt}
K_i = \left \{S \in K : \forall v \in S, f(v) \le i \right \}
\end{equation}
Intuitively, $f(v)$ represents the point at which $v$ enters the filtration and $\max_{v \in S} f(v)$ determines the point at which a simplex $S \in K$ enters the filtration.  Now we have the filtered complex $K_{f_{min}} \subseteq K_{f_{min} + 1} \subseteq \dots \subseteq K_{f_{max}} = K$.  Notice that if we reverse the inequality in Equation \ref{eq:imfilt}, we get an equally valid filtration, $K_{f_{max}} \subseteq K_{f_{max} - 1} \subseteq \dots \subseteq K_{f_{min}} = K$.  We will refer to these as the \emph{increasing} and \emph{decreasing} filtrations.

\begin{figure}[!ht]

\begin{subfigure}{\textwidth}
\center
\includegraphics[width=4.75in]{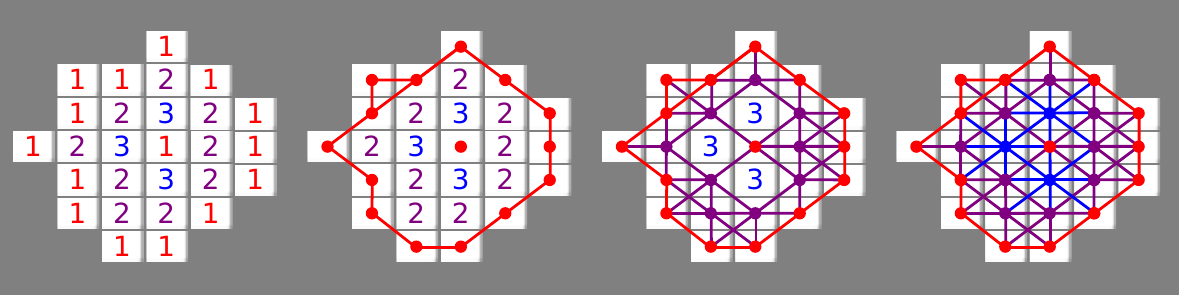}
\caption{Simple image with filtered complex}
\label{fig:toyim}
\end{subfigure}
\begin{subfigure}{\textwidth}
\center
\includegraphics[width=4.0in, trim=0in .5in 0in 0in, clip]{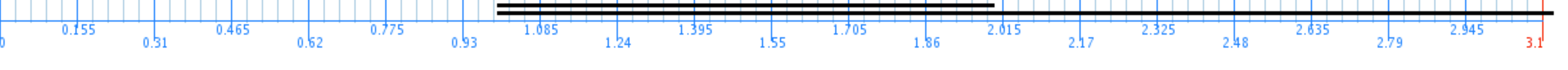}
\caption{$\beta_0$ barcode for above image}
\end{subfigure}
\begin{subfigure}{\textwidth}
\center
\includegraphics[width=4.0in, trim=0in .5in 0in 0in, clip]{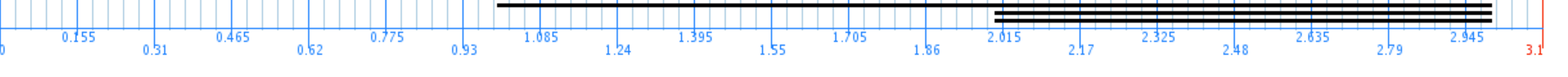}
\caption{$\beta_1$ barcode for above image}
\end{subfigure}
\caption{Constructing an increasing 1D-filtration on an image}
\label{fig:toy}
\end{figure}

As each vertex is associated with a pixel in the original image, it is natural use the pixel intensity (i.e., the grayscale value of each pixel) to assign a value to each vertex.  This forms the basis of what we will call the \emph{intensity filtration}.  A toy example of the increasing intensity filtration is shown in Figure \ref{fig:toyim}.  The colors represent the point in the filtration when the vertices and edges are added (we do not shade triangles for aesthetic reasons).  

We define an additional filtration by associating the distance from the lesion border, as given by the radiologist, to each pixel.  We call this the \emph{border filtration}.  The increasing border filtration produces an `annulus' which grows until it fills the lesion. The decreasing filtration produces a misshapen `disc' which expands from the center of the lesion.  While this is clearly not topologically useful for classification, in practice the combination of the border filtration with the intensity filtration gives better classification results than using the intensity filtration alone. 

To simplify the computational difficulties encountered with two-dimensional filtrations, we use one-dimensional filtration \emph{slices} to approximate the two-dimensional filtrations.  Let $K^b_{i}$ represent the border filtration and $K^p_{j}$ represent the pixel intensity filtration.  We divide the range of the border-filtration into 20 equally spaced slices.  At each slice $i$, we use the intensity filtration to compute the persistent homology of the subcomplex $K^b_{i}$.  This gives rise to the filtered complex $K_{if_{min}} = K^b_{i} \cap K^p_{f_{min}} \subseteq \dots \subseteq K_{if_{max}} = K^b_{i} \cap K^p_{f_{max}} = K^b_{i}$.

We can treat each of these one-dimensional barcodes as a different measurement on the lesion.  The options of the increasing or decreasing filtration on the border and intensity functions, as well as the $\beta_0$ and $\beta_1$ barcodes gives eight barcodes at each slice, yielding a total of 160 barcodes computed on each lesion.  

To account for differences in the pixel scaling, whether due to image formatting or differing CT scanners, we normalize the pixel range from zero to 1.  We stop infinite barcodes at 1.1 so that a differing number of infinite bars does not immediately separate two lesions (some hemangiomas, for example, have two dense regions while others have 3 or 4).  

\subsection{Feature Generation and Machine Learning}

To make use of existing machine learning techniques, it is necessary to provide a vector of measurements for each lesion.   Using the barcode distance, we can create a vector of relative measurements by computing the matching distance between each lesion and all other lesions (including itself).  In other words, we use the entire set of 132 images as the comparison set to generate our feature vector.  We do this even when restricting ourselves to a smaller subset of lesions.  This allows us to obtain information even from the lesion type sets that are too small for classification.  

Since we have 160 barcodes for each lesion, we choose to sum the 160 distances to create a vector of size 132 for each lesion.  This vector can then be used in traditional machine learning algorithms.  We choose to use an implementation of the support vector machine (SVM) called LibSVM to test the classification accuracy on various subsets of the data with the one-dimensional and two-dimensional filtrations \cite{libsvm}.

\section{Results}

\begin{figure}[!htb]
\begin{center}
\begin{subfigure}{\textwidth}
\begin{center}
\includegraphics[width=4.75in, trim=2.5in 0in 0.5in 0in]{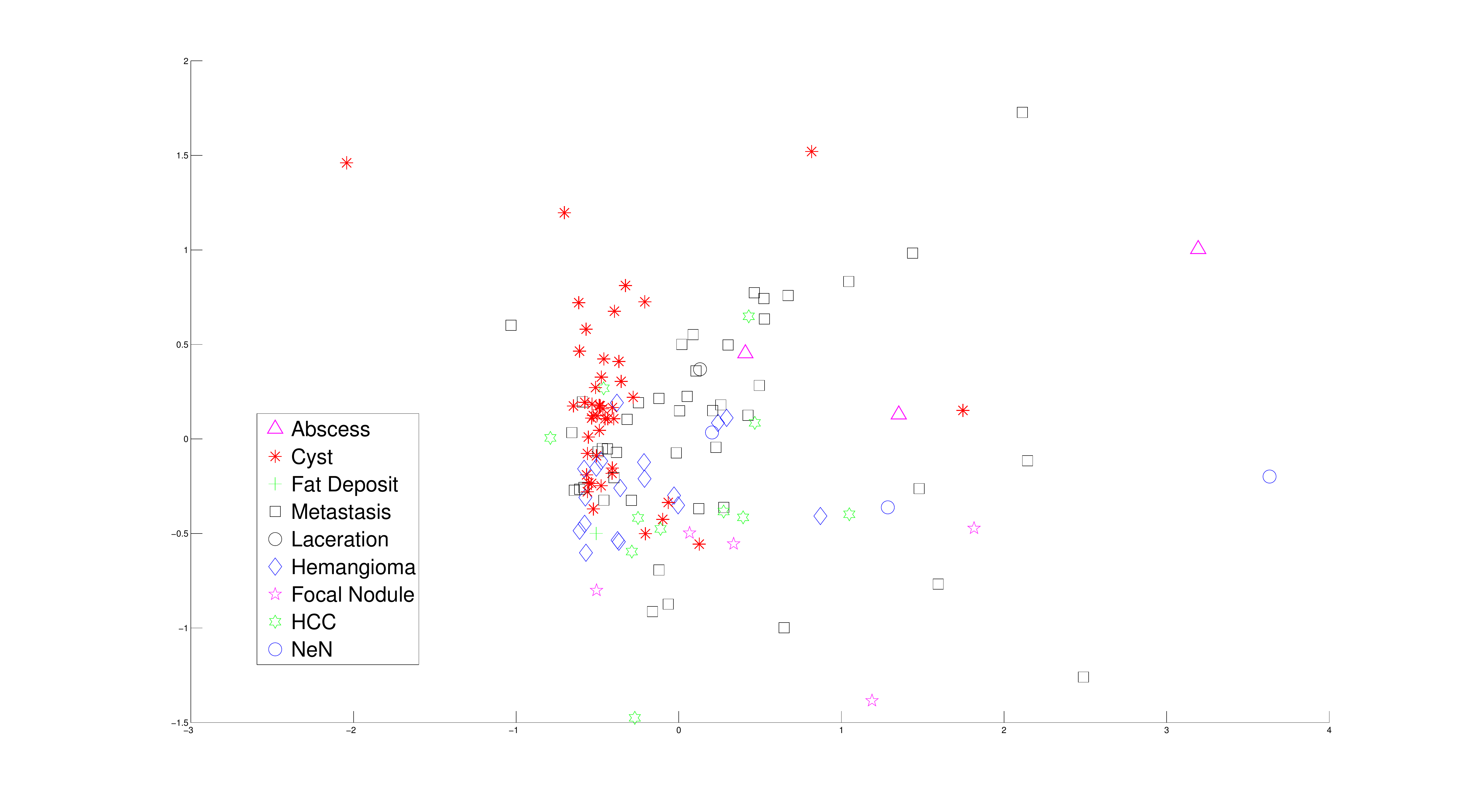}
\caption{A 2D View of 132 Hepatic Lesions}
\label{fig:cmds2}
\end{center}
\end{subfigure}

\begin{subfigure}{\textwidth}
\begin{center}
\includegraphics[trim=2.5in 0in 0.5in 0in, width=5in]{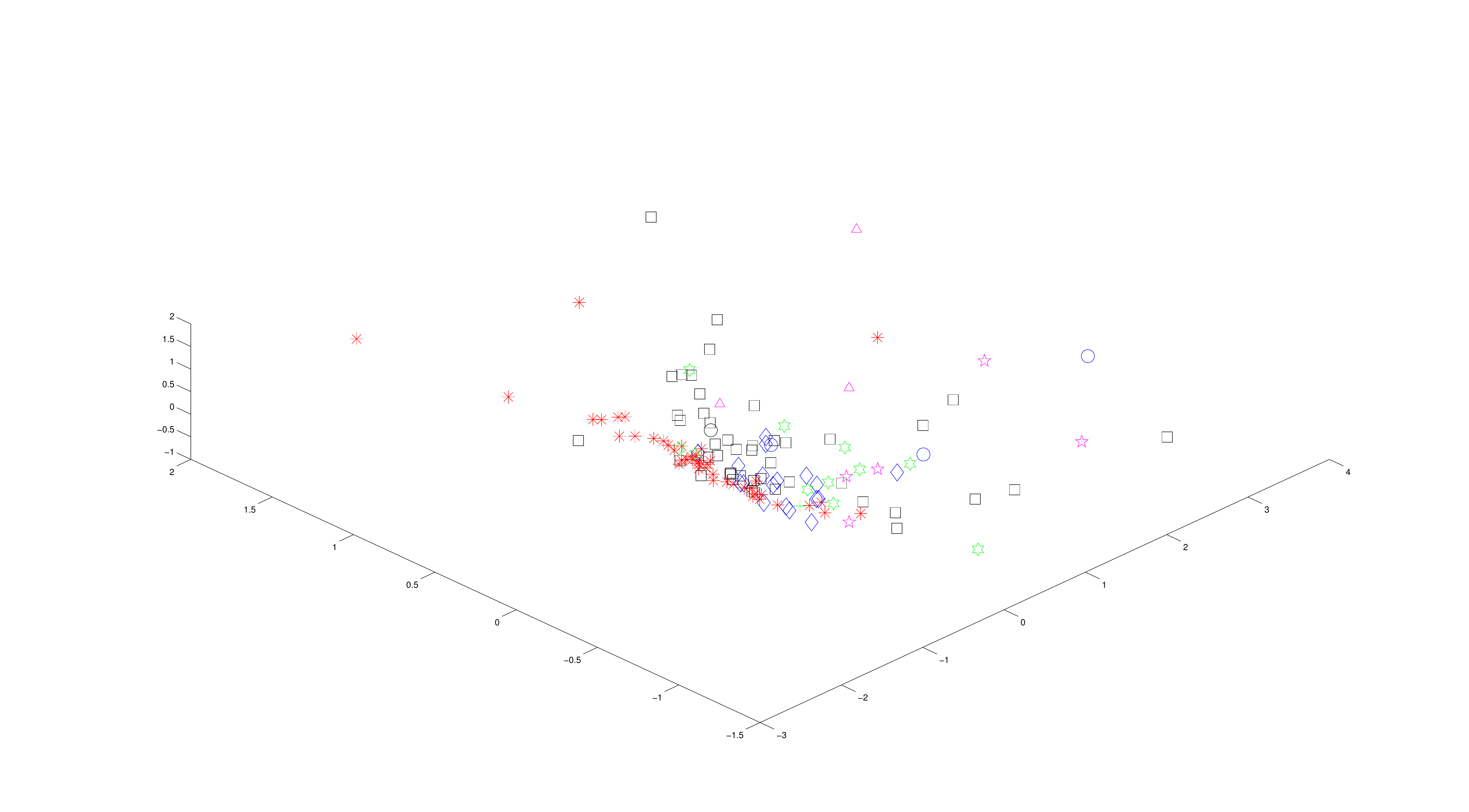}
\caption{A 3D View of 132 Hepatic Lesions}
\label{fig:cmds3}
\end{center}
\end{subfigure}
\caption{Visualizations of Topological Features}
\label{fig:viz}
\end{center}
\end{figure}

For the purpose of building intuition, we used classical multidimensional scaling (CMDS) on the distance matrix, using the above feature vectors as columns, to produce 2D and 3D visualizations of the lesions, shown in Figure \ref{fig:viz}.  Note that the axes simply give the coordinates of the embedding given by CMDS and that the vertical axis in Figure \ref{fig:cmds3} is pointed at the reader in Figure \ref{fig:cmds2}.

We use a SVM and leave one out cross-validation (LOOCV) to test the efficacy of our methods.  Because the dataset available to us is very unbalanced, see Section \ref{sec:data}, we present results from four different subsets of the data.  The first is the full dataset.  In the second subset (HcHeCM), we use the HCCs, the hemangiomas, the cysts, and the metastases.  In the third dataset (HeCM) we test on the hemangiomas, cysts, and metastases.  This data set is most comparable to the sets used in previous classification works \cite{autoretrieve}.  In the final dataset (CM) we remove the hemangiomas, leaving only the cysts and metastases.

\begin{table}[h]
\caption{SVM Classification Accuracies for 1D and 2D Filtrations}
\label{tbl:SVM} 
\begin{center}
\begin{tabular}[h]{| l | c | c | c | c |}
\hline
Filtration & Full  & HcHeCM & HeCM & CM \\ \hline
1D (Intensity) & 55.30\% & 59.66\% & 63.89\% & 80.00\% \\ \hline
2D & 66.67\% & 72.27 \% & 80.56\% & 85.56\% \\ \hline
\end{tabular}
\end{center}
\end{table}

We used the Gaussian kernel (also called the radial basis function), $e^{\sigma^2 |u-v|^2/2}$, in combination with the SVM.  We performed an exponential parameter sweep to find reasonable values for the parameter $\sigma$ and the SVM cost parameter $C$ for each data set.  In Table \ref{tbl:lesion_miss} we show the misclassification rates of each lesion type in the HeCM dataset.  

\begin{table}[h]
\caption{HeCM \% Classification Accuracy by Lesion Type}
\label{tbl:lesion_miss} 
\begin{center}
\begin{tabular}[h]{| c | c | c | c | c |}
\hline
Filtration & \% of HeCM & \% of Heman.  & \% of Cysts & \% of Metas. \\ \hline
1D & 63.89\% & 27.78\% & 77.78\% & 64.44\% \\ \hline
2D & 80.56\% & 72.22\% & 88.89\% & 75.56\% \\ \hline
\end{tabular}
\end{center}
\end{table}

Upon examination of the lesions which were misclassified, we noticed that many of the lesions were significantly larger than the median lesion area (1285.5 pixels) of the dataset.  Taking HeCM, we performed the same analysis as above, but removed lesions with various pixel areas.  The results are summarized in Table \ref{tbl:size_class}. 

\begin{table}[h]
\caption{Classification by Lesion Size of HeCM}
\label{tbl:size_class} 
\begin{center}
\begin{tabular}[h]{| c | c | c | c | c |}
\hline
Lesion Size by Area & \% Accu. & \# of Heman. & \# of Cysts & \# of Metas. \\ \hline
All & 80.56\% & 18 & 45 & 45 \\ \hline
$<$10000 px & 83.50\% & 18 & 42 & 43 \\ \hline
$<$5000 px  & 86.96\% & 16 & 39 & 37 \\ \hline
$<$2500 px  & 86.25\% & 14 & 32 & 34 \\ \hline
$<$1250 px  & 91.53\% & 8 & 28 & 23 \\ \hline
\end{tabular}
\end{center}
\end{table}

\section{Discussion} 
A large portion of the misclassifications are due to the difficulties in normalizing for the number of pixels present in the lesion.  A large lesion, has many more potential features than a small lesion.  Since the matching metric involves a matching of bars, a larger lesion will be a greater distance from a smaller lesion, even if the tissue structure is the same, because of the unmatched bars.  A method of accounting for the differences in lesion size would improve our results considerably.  In addition to this, our results are potentially sensitive to the lesions available for comparison.  A different comparison set (or even a synthetic comparison set) could improve or degrade the results.  Addressing both these issues are potential future directions of research.

Nevertheless, the current results are comparable to more traditional feature based classification methods \cite{autoretrieve, gabor}.  This demonstrates that multidimensional persistence is viable candidate for integration with other methods of developing features in radiological data.  In particular, our methods may be complementary to the standard techniques currently in use.

These results also demonstrate the power of combining topology and geometry via persistence.  In this case, it is clear that using a radial geometry (via the border filtration) significantly improves the classifying power of barcodes, especially in the case of hemangiomas, which are characterized by large homogenous regions on the periphery of the lesion.  This demonstrates one way in which this methodology captures radiological observations.  

This procedure is flexible enough to be used in a variety of contexts.  Any method of assigning values to each pixel could be used as a filtration to generate a barcode.  Filtering could be done on the image or on the barcodes (for example, removing bars of small length).  If a different geometric filter is called for in an application, it can be easily accommodated by our methods.

Additionally, the output from our algorithm was ready for input into an existing machine learning algorithm.  This demonstrates that our algorithm is easily integrated into existing computational machinery and can be combined with more traditional methods of feature generation.
 
\section{Conclusion}
We have implemented and tested a methodology for the classification of hepatic lesions.  Using multidimensional persistent homology and a support vector machine, we demonstrated the ability of multidimensional persistence to combine different features of interest for improved results over a one-dimensional filtration.  Our computational framework can be used in a variety of image classification problems outside of lesion classification and can be tailored to the specific application by changing the filtrations used on the image. We achieved comparable results to the traditional classification methods of hepatic lesions and, because our methods are topologically based, this makes them good candidates for integration with classical non-topologically based image processing techniques.

\section*{References}

\bibliographystyle{elsarticle-num}

\section*{Vitae}
{\bf Aaron Adcock} is a Ph.D candidate in the Department of Electrical Engineering at Stanford University and is advised by Gunnar Carlsson and Michael Mahoney.  His research interests include applications of computational topology, the analysis of complex networks, and data mining.  His current projects include developing new techniques in applying computational topology to medical images, integrating computational topology with existing machine learning techniques, and investigating the structure of complex networks using tree-like structures.

{\bf Daniel Rubin} is Assistant Professor in the Department of Radiology at Stanford University. Work in the lab lies at the intersection of biomedical informatics and imaging science. His NIH-funded research program focuses on developing informatics methods of knowledge representation, natural language processing, and decision support to improve the quality and consistency of radiology practice. Major projects include (1) developing methods to extract information and meaning from images for data mining, (2) developing statistical natural language processing methods to extract and summarize information in radiology reports and published articles, (3) resources to integrate images with related clinical and molecular data to discover novel image biomarkers of disease, and (4) translating these methods into practice by creating decision support applications that relate radiology findings to diagnoses and that will improve diagnostic accuracy and clinical effectiveness. 

{\bf Gunnar Carlsson} is the Swindells Professor of Mathematics at Stanford University and is a co-founder of Ayasdi Inc. Gunnar has a Ph.D. in Mathematics from Stanford University. In the last ten years, he has been involved in adapting topological techniques to data analysis, under NSF funding and as the lead PI on the DARPA ÒTopological Data AnalysisÓ project from 2005 to 2010. He is the lead organizer of the ATMCS conferences, and serves as an editor of several Mathematics journals. 
\end{document}